\documentclass[letterpaper]{article} 
\usepackage{aaai24}  
\usepackage{times}  
\usepackage{helvet}  
\usepackage{courier}  
\usepackage[hyphens]{url}  
\usepackage{graphicx} 
\usepackage{natbib}  
\usepackage{caption} 
\usepackage{algorithm}
\usepackage{algorithmic}

\usepackage{graphicx}
\usepackage{amsmath}
\usepackage{amssymb}
\usepackage{booktabs}
\usepackage{subfigure}
\usepackage{color}

\usepackage{newfloat}
\usepackage{listings}
\DeclareCaptionStyle{ruled}{labelfont=normalfont,labelsep=colon,strut=off} 
\floatstyle{ruled}
\newfloat{listing}{tb}{lst}{}
\floatname{listing}{Listing}
\title{Full-Body Motion Reconstruction with Sparse Sensing from Graph Perspective}
\author {
    Feiyu Yao\textsuperscript{\rm 1},
    Zongkai Wu\textsuperscript{\rm 2\correspondingauthor},
    Li Yi\textsuperscript{\rm 3,4,5\correspondingauthor}
}
\affiliations {
    \textsuperscript{\rm 1}2012 Lab, Huawei Technologies Co., Ltd\\
    \textsuperscript{\rm 2}Fancy Technology\\
    \textsuperscript{\rm 3}Tsinghua University\\
    \textsuperscript{\rm 4}Shanghai Artificial Intelligence Laboratory\\
    \textsuperscript{\rm 5}Shanghai Qi Zhi Institute\\
    yaofeiyu1@huawei.com, wuzongkai@fancy.tech, ericyi@mail.tsinghua.edu.cn
}

\usepackage{bibentry}

\begin{document}

\maketitle

\begin{abstract}
Estimating 3D full-body pose from sparse sensor data is a pivotal technique employed for the reconstruction of realistic human motions in Augmented Reality and Virtual Reality. However, translating sparse sensor signals into comprehensive human motion remains a challenge since the sparsely distributed sensors in common VR systems fail to capture the motion of full human body. In this paper, we use well-designed Body Pose Graph (BPG) to represent the human body and translate the challenge into a prediction problem of graph missing nodes. Then, we propose a novel full-body motion reconstruction framework based on BPG. To establish BPG, nodes are initially endowed with features extracted from sparse sensor signals. Features from identifiable joint nodes across diverse sensors are amalgamated and processed from both temporal and spatial perspectives. Temporal dynamics are captured using the Temporal Pyramid Structure, while spatial relations in joint movements inform the spatial attributes. The resultant features serve as the foundational elements of the BPG nodes. To further refine the BPG, node features are updated through a graph neural network that incorporates edge reflecting varying joint relations. Our method's effectiveness is evidenced by the attained state-of-the-art performance, particularly in lower body motion, outperforming other baseline methods. Additionally, an ablation study validates the efficacy of each module in our proposed framework.
\end{abstract}

\section{Introduction}
Continuously full-body motion reconstruction from sparse motion sensing is crucial for applications in Augmented Reality and Virtual Reality (AR/VR), which demands highly accurate human motion poses to render vivid avatars in the digital world and do interactions. Common VR systems are composed by head-mounted displays and handheld controllers. These devices can provides resourceful abundant upper body motion information, yet they are unable to provide corresponding lower body motion data. The significant sparsity inherent in known data distribution makes the generation of realistic full-body motion a particularly challenging endeavor for conventional methods based on human kinematics \cite{finalk} and matching motions \cite{CoolMoves}. 

Various learning-based methods have been made to generate full-body avatars from sparse inputs in AR/VR \cite{VAE} \cite{agrol} \cite{AvatarPose} \cite{AvatarPose}. These methods in these diverse studies essentially entail the extraction of features from sparse sensor data, devoid of considerations for human body joint relationships. Subsequently, these extracted features are integrated into various network architectures that similarly also lack a profound consideration of the interdependence among human body joints. The homogenization of these methodologies confines the development of reconstructing human motion from sparse inputs to the realm of network structure updates. Also, the absence of sufficient human body information contributes to a notable disparity between the reconstructed outcomes of the lower human body and actual motion dynamics. 

To solve the problems mentioned above, we consider the human body from graph perspective and propose BPG to represent full body. The task is then transformed to be predicting missing nodes in established BPG. Considering the limited information available on missing nodes, the BPG is initialized and updated referring to node properties. The first stage is processing node features. Position feature and angle feature are fused since they share different transformation law and distribution. 
Temporal Pyramid Structure is proposed on fusing frame-level and clip-level features to build temporal properties for feature representation.  
To model spatial properties, features of limb joints and trunk joints are generated separately referring to the human skeleton dynamic. The generated motion features are assigned to be initial features in BPG.
In Node Feature Updating stage, the nodes in BPG is updated referring to joint relations. We split the node relations into static skeleton relations, dynamic skeleton relations and latent relations. Then the node features in BPG are updated in Graph Convolution Network with expressive edges generated from node relations. 

\begin{figure*}
\begin{center}
\includegraphics[width=0.82\linewidth]{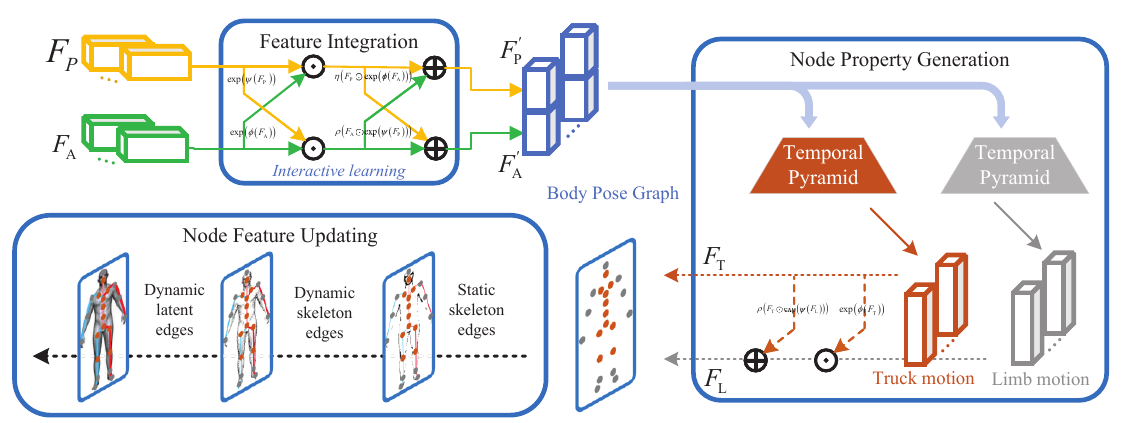}
\end{center}
   \caption{Illustration of our proposed structure. Inputs are sparse sensor position and rotational signals from VR system. Feature Integration module integrates position feature and rotation feature with different physical properties with interactive learning. 
   In Node Property Generation module, motion temporal property is achieved through Temporal Pyramid module. To gain motion spatial property, the limb motion feature is composed by trunk motion features and limb local motion features. The trunk and limb feature then serve as initial node features in Body Pose Graph. In Node Feature Updating, graph convolution network with different edges modeling different joint relations is applied to update nodes.}
\label{framwork}
\end{figure*}

Our main contributions are summarized as follows:

\begin{itemize}
    \item We are the first to conduct research on full body pose reconstruction with sparse sensing from graph perspective. The task is viewed as predicting missing nodes in an established graph.

    \item We propose a framework to reconstruct full body motions via Body Pose Graph (BPG). Motion features with temporal and spatial properties are generated and assigned to be initial node features. Then a Graph Neural Network with expressive edges is applied to updated nodes. Full body motion sequence is generated from the node-associated joint movements.
    
    \item Experiment results demonstrate that our framework achieves state of the art performance on  on full-body avatar estimation from sparse inputs. Further analysis shows the contribution of each component to the performance improvement, especially in the lower body joints.   

\end{itemize}

\section{Related Work}

\subsection{Full-Body Motion Reconstruction From Sparse Inputs}

Primary researches on this area make attempts on fully-body motion reconstruction from 6 IMU sensors on human body (head, arms, pelvis and legs). \cite{Sip} proposes a joint optimization framework based on statistic body models. \cite{Dip} applies learning method BiRNN with body models to do the estimation. \cite{Transpose} proposes a multi-stage learning based method where multiple subtasks and losses are designed to restrain pose generation. \cite{PIP} relies on physical models to refine the poses generated from learning methods.

However, requiring six IMUs is still excessive, as they are costly and logistically inconvenient to deploy. Reconstructing full body motion from common VR system will be more advantageous and flexible. \cite{CoolMoves} first utilizes sparser inputs from current consumer-grade VR systems (with headset and hand controllers) to estimate full-body motion. It makes full body motion reconstruction much easier. However, it estimates poses based on matching from a dataset with only 5 types of activities. It can hardly handle diverse activities out of dataset. \cite{Lobstr} proposes a Gated Recurrent Unit - based method to estimate lower-body pose while achieving upper-body with IK solver. It queries the confidence of lower-body motion reconstruction especially when upper-body and lower-body have weak correlations. Thus apart from sensor data from VR devices, it also requires a sensor on human waist. \cite{VAE} proposes a VAE(Variance AutoEncoders)-based method to generate poses from VR devices. However, it assumes the directions of pelvis in each frame should be the same. \cite{AvatarPose} proposes a Transformer-based structure to generate global orientation and local joint orientations. Orientations will then be input to body models to generate joint poses. 
In the domain of human body reconstruction utilizing only head-mounted devices, several methods have also been developed and explored. 
\cite{egoego} estimates full-body human motions from only egocentric video for diverse scenarios. \cite{QuestSim} proposes a reinforcement learning based framework and, together with physical simulator, can generate vivid leg motions even when the input is only the 6D transformations of the HMD. Despite these methods' promising performance, leveraging joint motion relations in the human body can likely yield better results. Hence, our work introduces this prior knowledge through a graph-based approach.


\subsection{Graph Neural Networks}

Graph Neural Networks \cite{gnn_node_prediction}\cite{gnn_edge_prediction} process data that can be represented as graph. Nodes representations will be iteratively updated by messages passed from their neighbors. Typical message passing methods include convolution-based methods \cite{gnn_node_prediction} and attention-based \cite{gat}. One research field related to sparsity is handling graphs with missing nodes. \cite{SAT} develops a distribution match based GNN Transformer-like method for attribute-missing graph. \cite{GCNMF} introduces Gaussian 
mixture model to represent missing data in Graph Convolutional Network. \cite{PaGNN} utilizes a partial message-passing method to transmit observed features 
observed features in GCN-based model. \cite{pmlr} handles missing features in graph by minimizing Dirichlet energy and leading to a diffusion-type differential equation on graph.

Although they handle the graphs missing data, however, the graphs that these methods focus on have three characteristics. First, the node features have great similitudes. Also, node relations tend to be qualitative and simple. And the downstream tasks (for example node prediction) do not rely much on the accurate quantity of node features. A typical example will be graph in bibliographic data. While full-body motion reconstruction from sparse inputs needs accurate quantity node features. The joints also have their own characteristics and the relations between them have motion meanings. The methods mentioned above are unsuitable for handling joint missing human joint graph.

\subsection{Graph Neural Networks in Human Pose Estimation}

While there are currently no methods solving full-body motion estimation from sparse inputs utilizing Graph Neural Networks (GNN), GNN, renowned for their heightened interpretability, have found widespread application in tasks associated with human pose. For example, action recognition \cite{stsgcn,Disentangling,shift,Decoupling,Multi-scale,lstm,Context,Actional,directed,Channel-Wise,Two-Stream,Revisiting,Multi-Stream,Spatio-Temporal} and 3D human pose estimation from 2D \cite{hyperbolic,hypergraph,Weight-Sharing,Non-local,Semantic,modulated,motion-prediction,global}.

\cite{stsgcn} is the first work that applies GNN in action recognition. After that,
various researches have been proposed on GNN-based method in action recognition. Some focus on improving the graph structure itself. For example, \cite{shift} proposes shift operations and lightweight point-wise convolutions to provide flexible receptive field for graph. \cite{lstm} integrates GNN with attention mechanism and lstm to increase representation ability. Some focus on empowering node relations to be more expressive. \cite{directed} generates graph edges with directions based on kinematics while \cite{Multi-Stream,Spatio-Temporal}  generates edges containing temporal information or spatial information by learning methods.

Different from action recognition tasks which focus on pose classification, 3D human pose estimation aims to reduce the estimation errors on all joints in pose. In order to solve this harder task, various methods focus on more powerful graph structure and more contextual edges. The more powerful graph includes graph in Non-euclidean space \cite{hyperbolic}, hypergraph neural networks \cite{hypergraph}, and so on. More advanced graph updating methods are also proposed, for example \cite{Weight-Sharing,Non-local,Semantic,modulated,stgcn}. Also, more human priors are utilized. For example, \cite{motion-prediction} establishes dynamic GNN based on human motion prediction. \cite{global} reveals the importance of decoupling global information from joints. \cite{hard_pose}, \cite{BMVC} analyzes the multi-hop relations between human graph nodes and models then in updating methods. 

The various methods mentioned above motivate us to introduce graph in fully-body pose estimation from sparse inputs. However, these methods mainly focus on tasks where sparsity hardly exists. Features for almost all joints are supplied as input. While our task only provides sensor data on 3 joints as inputs and expects accurate estimation in 22 joints. Above mentioned methods in related human pose methods can hardly be applied to our task. 


\section{Methods}

In this section, we first formalize the process of full body motion reconstruction with sparse sensing and understand the task from graph perspective. Then building process of BPG is introduced. After node initialization, BPG is updated referring to several joint relations and all joint motions are generated.

\subsection{Problem Formulation}

This work focuses on full-body motion reconstruction with measurements from one headset and two hand controllers, a common configuration in commercial VR device. The inputs are cartesian coordinates $\mathbf{p}^{1 \times 3}$ and orientations in axis-angle representation $\boldsymbol{\Phi}^{1 \times 3}$ of headset and hand controllers. The outputs are local rotation angles between joints and their parent joints $\theta$. considering real-time requirements in application scenarios, the issue is formalized an online problem:



\begin{equation}
\mathbf{\theta}_{\text{N}}^{\text{1:F}}=f\left(\left\{\mathbf{p}^{\mathbf{w}}, \boldsymbol{\Phi}^{\mathbf{w}}\right\}_{\text{(N-K):N}}^{\text{1:S}}\right),
\end{equation}

in which $S=3$ corresponds to the number of joints tracked by the VR system, $F=22$ is the number of joints used to represent the full-body motion. 
The movements of joints in current frame ($\text{N}$) are generated from sensor data in previous K frames ($\text{(N-K):N}$). The final full human body can be rendered from the outputs $\theta$  with human body model. 

From graph perspective, we view the full-body as a graph with 22 nodes. For N-th frame full-body motion reconstruction, motions of 3 nodes in graph are known and used. They are the positional and angular motions of head and two hands in previous K frames. Thus the task is transformed to be completion of the missing 19 nodes in graph. Feature Integration module and Node Property Generation module extract various features from movement sequences of three known nodes, assign features to different nodes and endow these node features with characteristics corresponding to the joints properties.

Feature Integration module integrates different sensor signals as normalization. Node Property Generation module generates features with temporal and spatial properties for nodes as initial value.  In Node Feature Updating module, the node features are updated by GCN with expressive edges.

\subsection{Node Feature Initialization}

Given the constraints of a limited number of known nodes and the valuable information associated with them, the sparse sensor data from sensors undergoes an abstraction process. This process leads to the extraction of features related to the whole body motions. Subsequently, these extracted features are assigned to all nodes, thus serving as the initialization for the graph structure.

\subsubsection*{Feature Integration}

The angular measurements and positional measurements have totally different distribution and follows different math laws for transformation, we propose Feature Integration module to fuse them. 

The measurements from each VR system device are joint position $\mathbf{p} \in R^{1 \times 3}$ and joint angular representation $\theta \in R^{1 \times 6}$ (elements in rotation matrix $R^{3 \times 3}$ first row and second row). We augment the features by differentiate the measurements and get joint velocity $\mathbf{v} \in v^{1 \times 3}$ and joint angular velocity representation $\omega_t \in R^{1 \times 6}$(elements in rotation velocity matrix $R_{t-1}^{-1} R_t$ first row and second row ). Joint position and joint velocity compose translation feature $F_{\text {$P_{sensor}$ }}$. Joint angular representation and joint angular velocity representation  compose joint rotation feature $F_{\text {$A_{sensor}$  }}$.

\begin{equation}
F_{\text {$P_{sensor}$ }}^{S \times 6} = \begin{bmatrix}
(\mathbf{p}_t^1, \mathbf{v}_t^1)^{1 \times 6} \\
\ldots\\

(\mathbf{p}_t^s, \mathbf{v}_t^s)^{1 \times 6} \\
\end{bmatrix}
\end{equation}

\begin{equation}
F_{\text {$A_{sensor}$  }}^{S \times 12} = \begin{bmatrix}
(\theta_t^1, \omega_t^1)^{1 \times 12} \\
\ldots\\
(\theta_t^s, \omega_t^s)^{1 \times 12} \\
\end{bmatrix}
\end{equation}






$F_{\text {$P_{sensor}$ }}$ and $F_{\text {$A_{sensor}$  }}$ describes the joint motion from different perspective and shares totally different geometric properties. Thus, to enhance the feature representation ability and eliminate feature geometric differences, we utilize dual interactive learning to generate new feature referring to \cite{graph_inter}.

\begin{align}
F_{\text {P }}^{\prime}&=F_{\text {P }} \odot \exp \left(\phi\left(F_{\text {A}}\right)\right) - \rho\left(F_{\text {A}} \odot \exp \left(\psi\left(F_{\text {P }}\right)\right)
\right)  \notag \\
 \quad F_{\text {A}}^{\prime}&=F_{\text {A}} \odot \exp \left(\psi\left(F_{\text {P }}\right)\right) + \eta\left(F_{\text {P }} \odot \exp \left(\phi\left(F_{\text {A}}\right)\right)\right)
\end{align}
$\psi$, $\rho$ are 1D convolutional layers. $exp$ is used to map different features onto similar distribution spaces. The Feature Integration module is applied to  integrate the motion information of nodes and output fused node features that incorporate both positional and angular information.


\subsubsection*{Node Temporal Motion Property Generation}

As stated in the Problem Formulation, the framework's input comprises motion information from $k$ preceding frames, while the output entails the motion of the current frame. Thus this module is designed to mitigate temporal disparities in features originating from different frames. Also, 
As highlighted in \cite{motion_law}, motion continuity stands out as a distinct characteristic of human motion. Motion details within each frame can be inferred from the surrounding contextual frames (clip). In this section, we apply the modeling of motion continuity as a guidance to mitigate temporal disparities.

\begin{figure}
\begin{center}
\includegraphics[width=1.02\linewidth]{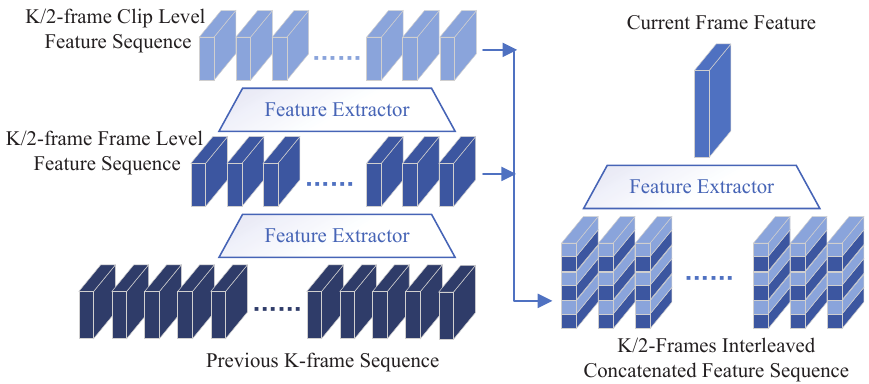}
\end{center}
\caption{Temporal Pyramid Structure}
\label{pyramid}
\end{figure}


For node temporal properties, to better capture the joint motion temporal properties, we design Temporal Pyramid Structure in Figure \ref{pyramid}. The inputs are $k$ previous frame features and the outputs are high dimensional temporal features of current frame. The feature extractor is based on SCI-Block \cite{SCINet}, which is a CNN-based time series model with output dimension adjustable. In Temporal Pyramid Structure, three Feature extractor are applied. First one and second one extract frame level and clip level features. The two level features are concatenated in an interleaved manner. The third extractor is applied to generate motion features for current frame.


\subsubsection*{Node Spatial Motion Property Generation}




Human motions in different joints have different spatial properties.
\cite{trunk_limb_1} claims the important impact of trunk on human body motion and reveals the different property of trunk and limb joints. 
As the human skeleton kinematic \cite{directed} \cite{directed_1} reveals, joints farther from the center of the human body are always physically controlled by an adjacent joint which is closer to the
center. In this context, limb joints act as child joints relative to trunk joints, resulting in limb joint motions being composed of both local limb joint movements and trunk joint motions. To address the challenge of predicting joints located distantly from the body's center and to capture this directed control relationship, we propose a unidirectional interactive learning approach. This method guides the extraction of limb motion features by leveraging the guidance from trunk motion features. This module's mechanism is described as followed.



\begin{align}
F_{\text{L}}=F_{\text{L}} \odot \exp \left(\phi\left(F_{\text{T}}\right)\right) + \rho\left(F_{\text{T}} \odot \exp \left(\psi\left(F_{\text{L}}\right)\right)
\right)  \notag \\
\end{align}

\begin{figure}
\begin{center}
\includegraphics[width=0.4\linewidth]{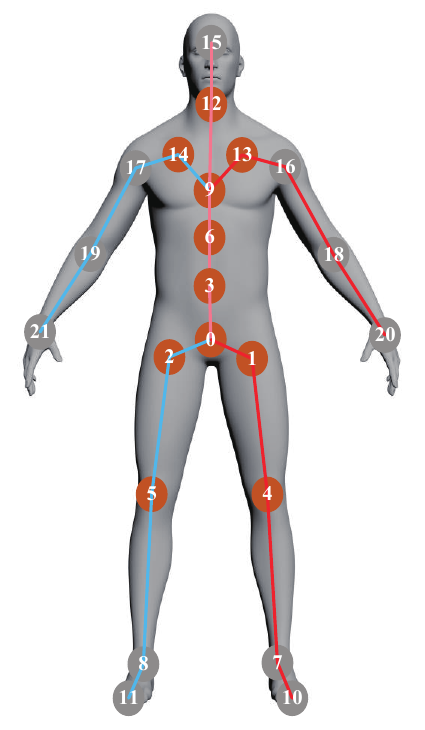}
\end{center}
\caption{Index of human body joints}
\label{learned edge}
\end{figure}

 $F_{T}$ and  $F_{L}$ are trunk motion features and limb motion features generated by temporal pyramid separately.
$\phi$ and $\psi$  are convolutional networks for generating sub-structure level features.
The interactive learning mechanism used here is similar to Feature Integration module.
The trunk and limb motion features generated are then assigned to corresponding trunk nodes (joint 0,1,2,3,4,5,6,9,12,13,14 in Figure \ref{learned edge}) and limb nodes (joint 7,8,10,11,15,16,17,18,19,20,21 in Figure \ref{learned edge}) as initial features.



\subsection{Node Feature Updating}

Node Feature Updating aims to capture diverse joint relationships through a Graph Convolutional Network with expressive edges. In Section 1, we initially revisit the vanilla Graph Convolutional Network, updating joint features considering the static human skeleton as edges. Subsequently, Section 2 introduces Node Updating using a graph convolutional network with expressive edges.


\subsubsection*{Vanilla GCN}

We will first review the vanilla graph convolution network. 
Given a graph $\mathcal{G}=(\mathcal{V}, \mathcal{E})$, it consists of the nodes $\mathcal{V}$ and the edges $\mathcal{E}$. We revisit a generic GCN layer defined as follows:

\begin{equation}
\mathbf{X^{'}}=\sigma({\mathbf{W}} \mathbf{X A})
\end{equation}

where $\mathbf{A} \in \mathbb{R}^{N \times N}$ is an adjacency matrix with $N$ nodes, indicating the connections between nodes. In 2D-3D human pose estimation, a task predicts 3d coordinates from pictures, the adjacency matrix is often established referring to human skeleton. If there is a bone connection between joint $j$ and the joint $i$, then $a_{i j}=1$. Otherwise, the value will be set to zero $a_{i j}=0$. We denote the input node features as $\mathbf{X} \in \mathbb{R}^{N \times C_{i n}}$. Each node corresponding to a $C_{i n}$-dimensional feature vector.  The learnable weight matrix $\mathbf{W} \in \mathbb{R}^{C_{i n} \times C_{\text {out }}}$ is set to adjust feature's dimension to be expected. The $\sigma(\cdot)$ is common activation function.

Consider i-th node, the node feature of node i is $X_i$. Corresponding adjacency matrix slice will be $A_i$, with j-th element being $a_{ij}$. $S_i$ represents the joints have bone connections with joint i. $a_{ij} = 1$ if $j \in S_i$ and $a_{ij} = 0$ if $j \not \in S_i$. The update of i-th node in vanilla GCN is expressed as: 

\begin{equation}
\mathbf{X}_i^{\prime}=\sigma\left(\sum_{j \in S_i} \mathbf{W} X_j    a_{i j}\right)
\end{equation}

Although lots of improvements has been made in past, the Graph updating method updates each node synchronously, which assumes that useful information is evenly distributed and the confidence of joint features is at the same level.

\subsubsection*{Node Updating With Expressive Edges}

Vanilla graph convolution network has limited representation ability. It assumes that features in each node is reliable and node can be represented well by updates referring to constant graph edges built by human skeletons. Also, other strong hidden relationships among joint nodes exist and change with actions, these relations can hardly be modeled by vanilla GCN. For example, when the human is running, there is a strong relation between hand joint and foot joint. But when the human is sitting, there is no such strong relation. Considering above two limitations, we proposed GCN with multiple kinds of edge learned. To be specific, the edges in graph are dynamic corresponding to the current joint state instead of being constant. When human action changes, the edges can change simultaneously to better represent the node relations.

In our task, edges are represented as an adjacency matrix $\mathbf{A}^h \in \mathbb{R}^{22 \times 22}$.

\begin{equation}
\mathbf{A}^h=\mathbf{A}^{s}+\mathbf{A}^{l}
\label{edge_1}
\end{equation}

\begin{equation}
\mathbf{A}^s=\mathbf{A}^{s s}+\mathbf{A}^{d s}
\label{edge_2}
\end{equation}

$\mathbf{A}^{s} \in \mathbb{R}^{22 \times 22}$ is skeleton relation adjacency matrix, it describes the relations exist in human skeleton (to be specific, all the edges drawn in Figure \ref{learned edge}). $\mathbf{A}^{l} \in \mathbb{R}^{22 \times 22}$ is latent relation adjacency matrix, it describes potential links between nodes (node links not exist in Figure \ref{learned edge} ). 
Static skeleton relation adjacency matrix  $\mathbf{A}^{s s} \in \mathbb{R}^{22 \times 22}$ is built referring to the human skeleton of SMPL model. Joints connected in human skeleton will have edges with non-zero constant value in the corresponding place. Dynamic skeleton relation adjacency matrix $\mathbf{A}^{ds} \in \mathbb{R}^{22 \times 22}$ models the relations among joints in skeleton. It is also built referring to human skeleton of SMPL. However, the value of the edges will be determined by the features of nodes in graph. 
The values in $\mathbf{A}^{ds}$ and  $\mathbf{A}^{l}$ are learned by MLP structure seperately.





\begin{equation}
A = W_1 \phi (W_0 X + B_0 ) + B_1
\label{mlp}
\end{equation}

in which, $X \in  \mathbb{R}^{b \times nf}$, $W_0 \in \mathbb{R}^{h \times nf}$, $B_0 \in \mathbb{R}^{h}$, $W_1 \in \mathbb{R}^{o \times h}$, $B_1 \in \mathbb{R}^{o}$. $X$ is joint feature. $b$ is batch size, $n$ is number of nodes and $f$ is the dimension of feature. $h$ is the dimension of the hidden layer. $o$ is the dimension of output. $\phi$ is the ReLU activation function. 

The nodes are updated with above adjacency matrix. The final output of BPG are the axis-angle of each joint, which, togther with SMPL human model \cite{smpl-h}, will be referred to generate the position of each joint .

\subsection{Training and Loss}

The loss function is composed of rotational loss, positional loss and bone symmetric loss. 

\begin{equation}
L_{\text{final}} = L_{\text{rot}} + L_{\text{pos}} + L_{\text{bone}}
\end{equation}

$L_{\text{rot}}$ is absolute error loss on all joint axis-angles. $L_{\text{pos}} $ is absolute error loss on all joint positions. The accuracy of axis-angle and position of each joint is both crucial for full body reconstruction. \cite{AvatarPose}. $L_{\text{bone}}$ is human skeleton symmetric loss. It emphasises the relative position relations among joints and introduce human body priors for optimization.




\begin{equation}
L_{bone} = \sum_{i_{l},j_{l},i_{r},j_{r}} (||\hat{Y}_{i_l}^{pos} - \hat{Y}_{j_l}^{pos}|| - ||\hat{Y}_{i_r}^{pos} - \hat{Y}_{j_r}^{pos}||)
\end{equation}

Here, $\hat{Y}_{i}^{pos}$ represents the  predicted position of joint i. $(i_l, j_l) \in set_{r}$ are right human skeleton bones shown as blue lines in Figure \ref{learned edge}, 
$(i_r, j_r) \in set_{l}$ are left human skeleton bones shown as red lines in Figure \ref{learned edge}.








\begin{table}[t]
\fontsize{10pt}{10pt}\selectfont
\resizebox*{\linewidth}{!}{
\begin{tabular}{lrr|r}
\multicolumn{4}{c}{} \\
\hline Methods & MPJRE & MPJPE &  MPJVE \\
\hline Final IK & $16.77$ & $18.09$ & $59.24$ \\
CoolMoves & $5.20$ & $7.83$ & $100.54$ \\
LoBSTr & $10.69$ & $9.02$ & $44.97$ \\
AvatarPoser & $3.21$ & $4.18$ & $29.40$ \\
\hline 
AvatarPoser *& $3.01$ & $4.11$ & $27.79$ \\
Our method * & $\mathbf{2 . 49}$ & $\mathbf{3 . 3 4}$ & $\mathbf{2 2 . 8 4}$ \\
\hline
\end{tabular}
}
\caption{Performance Comparison among our method and baselines in AMASS dataset. Notice that * means the results are trained in our machine.}
\label{overall} 
\end{table}

\section{Experiemnt}

\subsection{Data Preparation and Evaluation Metrics}

CMU \cite{CMU}, BMLrub \cite{Bio} and HDM05 \cite{MP} in AMASS \cite{amass} dataset are employed . The datasets are randomly partitioned into training and testing subsets, comprising $90\%$ and $10\%$ of the data respectively, following the same setting as \cite{AvatarPose}.  

The metrics utilized for overall performance comparison are MPJRE (Mean Per Joint Rotation Error $\left[^{\circ}\right]$ ), MPJPE (Mean Per Joint Position Error $[\mathrm{cm}]$ ), and MPJVE (Mean Per Joint Velocity Error $[\mathrm{cm} / \mathrm{s}]$ ). In ablation study, to reveal the effect of each component on motion reconstruction, we list the estimated position error on each lower body joint.

\subsection{Performance Comparison With Baseline Method}

We compare our method with baseline methods in Table \ref{overall}. To be specific, there are Final IK, CoolMoves\cite{CoolMoves}, LoBSTr \cite{Lobstr}, VAE-HMD \cite{VAE}, AvatarPoser \cite{AvatarPose}. The results are referring to \cite{AvatarPose}. To be fair, we retrained AvatarPoser on our platform. Our method attains superior results across all three metrics, outperforming all other methods. By representing human body as Graph and modeling spatial-temporal relations among joints, our method surpasses than baseline method, notably in predicting unseen lower body joints, as illustrated in Table \ref{PJPE}.

\subsection{Performance Comparison With Offline Method}

Offline methods refer to methods outputting n length human pose sequence instead single frame in each inference. AGRoL \cite{agrol} is the state-of-art Offline method. In our method, we use 41 frame sensor sequence as input and ouput 1 frame in each inference. In Table \ref{Generative}, $\mathrm{AGRoL}_{41}$ represents that the lengths of input sequence and output sequence are both 41. Thanks to the feature generation method and graph based architecture, which are special designed for human body, our method performs better than AGRoL in all criteria in same condition. 
When extending the output sequence length to 192, AGRoL demonstrates commendable performance in MPJVE metric. However, this enhancement of the MPJVE metrics did not translate into superior results for the MPJRE and MPJPE metrics, which are more important for the restructure task. 
In contrast, our proposed methodology exhibits superior performance in both the MPJRE and MPJPE metrics, further substantiating its efficacy.

\begin{table}[t]
\fontsize{10pt}{10pt}\selectfont
\resizebox*{\linewidth}{!}{
\begin{tabular}{lrr|r}
\multicolumn{4}{c}{} \\
\hline Methods & MPJRE & MPJPE &  MPJVE \\
\hline 
VAE-HMD & $4.11$ & $6.83$ & $37.99$ \\
AGRoL $_{41}$  &  $2.59$  &  $3.64$  &  $23.24$   \\
AGRoL $_{196}$  &  $2.66$  &  $3.71$  &  $\mathbf{18.59}$   \\
\hline
Our method & $\mathbf{2.49}$ & $\mathbf{3.34}$ & $22.84$ \\
\hline
\end{tabular}
}
\caption{Performance Comparison with offline methods}
\label{Generative} 
\end{table}


\subsection{Ablation Study}


\begin{figure*}[!tbh]
\begin{center}
   \includegraphics[width=1\linewidth]{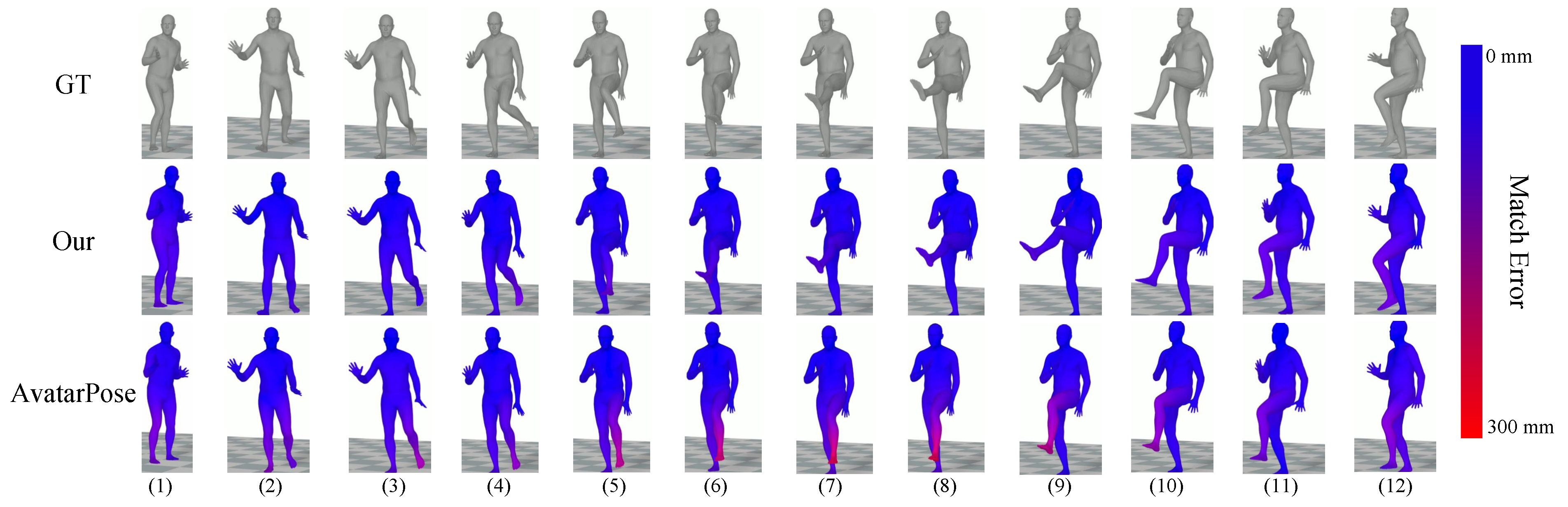}
\end{center}
   \caption{
   Visualization of estimated poses on an avatar involves a series of frames portraying a human front kick action. It encompasses three rows: the top row showcases avatars with ground truth (GT) poses, while the subsequent two rows display avatars generated by our approach and AvatarPoser. These avatars are color-coded to denote errors in each mesh.
   }
\label{visualization}
\end{figure*}

\begin{figure}[t]
\begin{center}
   \includegraphics[width=0.45\linewidth]{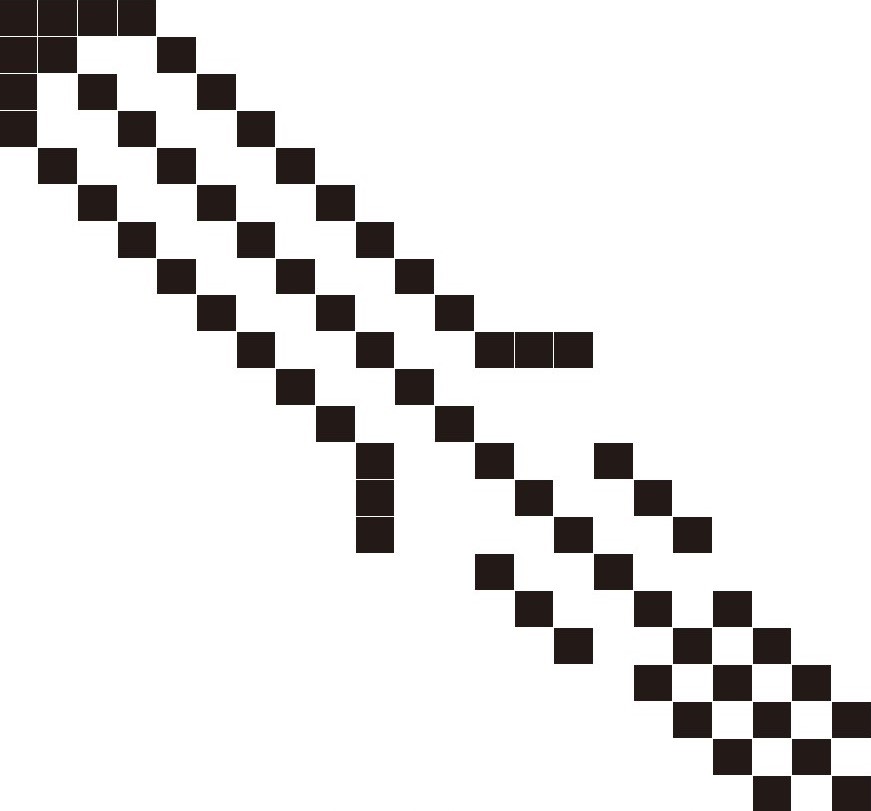}
   \quad
   \includegraphics[width=0.45\linewidth]{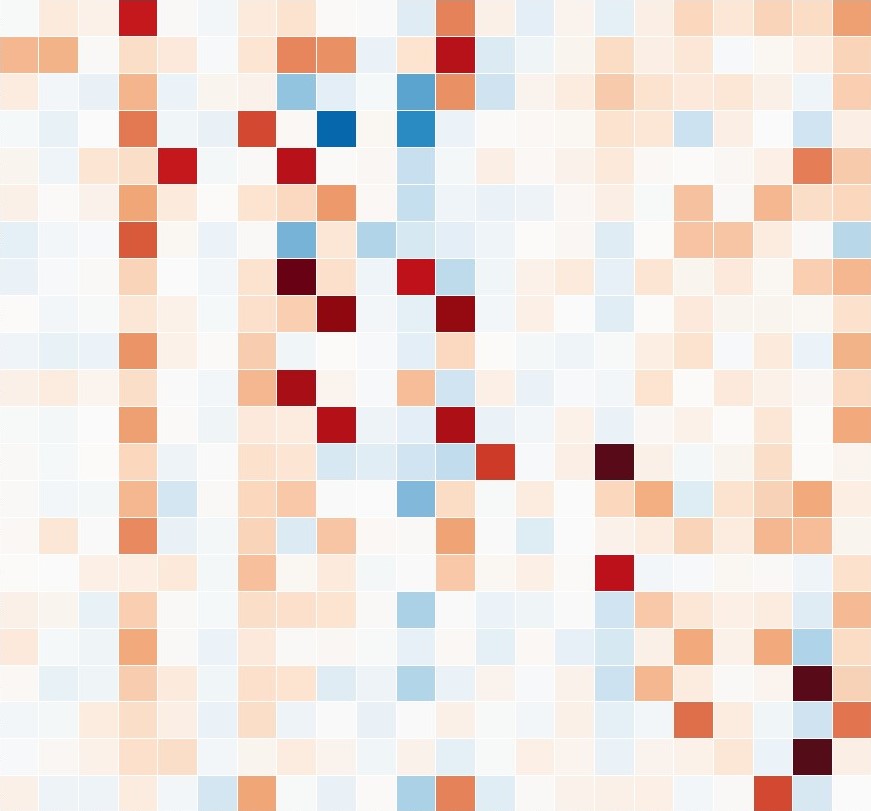}
\end{center}
   \caption{
   Left diagram depicts a 0-1 adjacency matrix representation of the skeletal connectivity within the human body. Conversely, right diagram showcases an adjacency matrix generated by the GCN with expressive edges. The deeper the color, the stronger the relationship between the nodes. Red indicates positive correlation, while blue indicates negative correlation.}
\label{fig:long}
\label{learned edge tsne}
\end{figure}

To dissect individual component functions, we conduct ablation studies across various cases. Findings are outlined in Table \ref{ablation}. Given our method's targeted focus on mitigating substantial estimation errors in lower body joints, we employ MPJPE-lower-body (MPJPE on joints 1, 2, 4, 5, 7, 8, 10, 11) to directly exemplify performance.

\begin{itemize}
    \item No Bone Symmetric Loss: The bone symmetric loss is not utilized in the framework.
    \item No Spatial Property: The node features are not generated separately for trunk joints and limb joints and the relations between trunk and limb is not considered.
    \item No Temporal Property: The temporal pyramid structure is replaced by  temporal feature extractor.
    \item No Feature Initialization: The Feature Integration process is replaced by simple MLP structure. 
    \item Vanilla GCN: 
    The nodes in BPG are updated through Vanilla GCN instead of the GCN with expressive edges.  

\end{itemize}  

As evident from Table \ref{ablation}, the absence of modules induces notable performance declines, particularly in the lower body region. This proves the efficacy of each component.

\begin{table}[!t]
\fontsize{10pt}{10pt}\selectfont
\resizebox*{\linewidth}{!}{
\begin{tabular}{lrrr}
\multicolumn{4}{c}{ } \\
\hline
Index  & AvatarPoser & Our method  & Improvement \\ \hline

Joint 1 &3.8 &3.1   & 18.84\%   \\
Joint 2  &3.8&3.2   & 15.79\%   \\

Joint 4 &6.9 &5.5   & 20.29\%   \\
Joint 5  &6.9 &5.5   & 20.29\%   \\
Joint 7  &10.1 &7.9   & 21.78\% \\
Joint 8  &10.1 &8.0  & 20.80\%   \\
Joint 10  &10.8 &8.4   & 22.22\%   \\
Joint 11  &11.0 &8.7  & 20.91\%     \\
All Joints & 4.11 &3.34 & 18.73\%   \\

\hline
\end{tabular}
}
\caption{Joint Position Error performance comparison between our method and AvatarPoser in lower body joints  [cm] }
\label{PJPE}
\end{table}

\begin{table}[!t]
\fontsize{10pt}{10pt}\selectfont
\resizebox*{\linewidth}{!}{
\begin{tabular}{lrr}
\multicolumn{3}{c}{ } \\
\hline Configuration & MPJPE & MPJPE-lower-body\\
\hline No Bone Symmetric Loss & $3.53$ & $6.75$ \\
No Spatial Property & $3.60$ & $6.88$ \\
No Temporal Property & $3.71$   & $7.10$ \\
No Feature Initialization & $3.53$ & $6.80$ \\
Vanilla GCN & $3.58$ & $6.88$\\
Default  & $3.34$ & $6.29$ \\

\hline
\end{tabular}
}
\caption{Ablation study}
\label{ablation}
\end{table}

\subsection{Visualization of Estimated Pose on Avatar}

In order to better analyze the estimation performance, we visualize the estimated poses on the whole avatar in figure \ref{visualization}. Each mesh triangle in avatar is rendered referring to the error of each estimated mesh vertex. Red represents large mesh vertex estimation error. The avatars in the first row show the ground truth poses. The avatars in second row and the third row are generated by our proposed method and baseline. As can be seen, the avatar generated by our method accomplishes the whole process of lifting and lowering the leg with little mesh error while the one generated by AvatarPoser accomplishes the action with errors and stiffness. Especially in frames (5) (6) (7) (8) (9), avatars generated by AvatarPoser can hardly even raise left leg as high as the ground truth.

\subsection{Analysis of Expressive Edges}

As shown in Figure \ref{learned edge tsne}, the adjacency matrix generated by GCN with expressive edges (right-hand figure) shows more joint relations than the static 0-1 adjacency matrix generated from the quantification of the human skeletal structure (left-hand figure). This indicated that, benefiting from the potent expressive capabilities of of GCN with expressive edges, our approach has yielded more comprehensive joint relationships in than human skeleton.

\section{Conclusion}

In this study, we approach the task of full-body motion reconstruction from sparse sensor input through a graph-based perspective, introducing the Body Pose Graph to represent the human body. In Node Feature Initialization step, different kind of VR system device features are first integrated. The new generated features are then processed to achieve spatial properties and temporal properties of joint motions before serving as initial node features in Body Pose Graph. Temporal property is generated by Temporal Pyramid Structure and Spatial property is generated referring to joint motion spatial relations. In the Node Feature Updating stage, we employ GNN with expressive edges to update node features within the Body Pose Graph. Our approach demonstrates exceptional estimation performance as evidenced by comprehensive evaluations.
Ablation studies validate the effectiveness of individual components. 
Visualizations of learned edges and estimated poses on avatars provide insights into learned motion relationships and our method's prowess in mesh-scale representations.

\bibliography{aaai24}

\end{document}